# Computing SHAP Efficiently Using Model Structure Information[1]

Linwei Hu, Ke Wang
Corporate Model Risk, Wells Fargo

**Abstract**

SHAP (SHapley Additive exPlanations) has become a popular method to attribute the prediction of a machine learning model on an input to its features. One main challenge of SHAP is the computation time. An exact computation of Shapley values requires exponential time complexity. Therefore, many approximation methods are proposed in the literature. In this paper, we propose methods that can compute SHAP exactly in polynomial time or even faster for SHAP definitions that satisfy our additivity and dummy assumptions (eg, kernal SHAP and baseline SHAP). We develop different strategies for models with different levels of model structure information: known functional decomposition, known order of model (defined as highest order of interaction in the model), or unknown order. For the first case, we demonstrate an additive property and a way to compute SHAP from the lower-order functional components. For the second case, we derive formulas that can compute SHAP in polynomial time. Both methods yield **exact** SHAP results. Finally, if even the order of model is unknown, we propose an iterative way to approximate Shapley values. The three methods we propose are computationally efficient when the order of model is not high which is typically the case in practice. We compare with sampling approach proposed in Castor & Gomez (2008) using simulation studies to demonstrate the efficacy of our proposed methods.

## 1  Introduction

Explaining a model's output is extremely important in many fields. In consumer lending, banks are required to provide reasons for declined credit application. In health care field, interpretation of predictions can help researchers better understand diseases. Among all interpretation methods, local interpretation explains individual predictions (Molnar, 2022). SHAP (SHapley Additive exPlanations), proposed in Lundberg & Lee (2017), is a popular local interpretation approach to attribute the prediction of a machine-learning model on an input to its features. The SHAP explanation method computes Shapley values (Shapley, 1953), a concept from coalitional game theory. The feature values of a data instance act as players in a coalition. Shapley value provides a way to fairly distribute the payout i.e., the prediction, among the features. The Shapley value is the only attribution method that satisfies the desirable properties efficiency, symmetry**,** dummy, and additivity**,** which together are considered a definition of a fair payout. There are different variations of SHAP for different models and applications (Lundberg & Lee, 2017,  Sundararajan & Najmi, 2020), examples include kernel SHAP and Baseline-SHAP (B-SHAP) and we will introduce them in more details in later sections.

Despite the benefits, computing Shapley values is extremely computationally expensive because it requires the order of $2^p$ function evaluations, $p$ being the number of features in the model. There are multiple ways aiming at speeding up computing Shapley values. We briefly introduce them in this section and will include more details in later sections. One strategy is that instead of computing Shapley values exactly, we can compute approximated Shapley values using sampling methods. Castro & Gomez (2008) and Štrumbelj & Kononenko (2014) proposed such estimates and was implemented in Captum[2]. We show in later sections that under some cases, these estimates can be far from the true values.

---

[1] The views expressed in the paper are those of the authors and do not represent the view of Wells Fargo.

To avoid the errors in sampling methods, we seek ways that can compute Shapley values precisely. Our main contribution is that we propose strategies of computing Shapley values exactly for a class of SHAP definitions that satisfy our additivity and dummy assumptions (see Assumption 1), by taking advantage of the model structure information. First, we show that Shapley values can be computed much more efficiently if the models have known structure, e.g., $f(\pmb{x}) = \sum_{v \subseteq \{1,2,...,p\}} f_v(\pmb{x}_v)$ and all the components are low dimensional ($|v|$ is small). In particular, main-effect plus two-way interaction models, $f(\pmb{x}) = \sum_i f_i(x_i) + \sum_{i,j} f_{ij}(x_i, x_j)$, have become popular recently due to the good model performance and interpretability (Lou, Caruana, Gehrke, & Hooker, 2013, Yang, Zhang, & Sudjianto, 2021, Hu, Chen, & Nair, 2022). For a black-box model, we can obtain these structures by using Functional ANOVA (fANOVA) decomposition (Hooker, 2004), a way to express complex models as the sum of components with low orders. Since the Shapley values of each functional component can be computed efficiently, by using the additive property of computing Shapley values introduced in this work later, the Shapley values of the complicated model can then be computed by adding the Shapley values of each low-order component.

Second, when we do not have a functional decomposition of a model, but we know the order of the model, we derive formulas that can compute Shapley values efficiently in polynomial time. Here, the order of a model is defined as the maximum order of interactions in the model. For example, an xgboost model with max_depth=4 has an order of 4, because it has at most 4-way interactions.

Finally, when the order of the model is unknown, it has been observed that the true underlying model is usually either low-order or approximately low-order, i.e., the high-order interactions are weak (Lou, Caruana, Gehrke, & Hooker, 2013, Yang, Zhang, & Sudjianto, 2021, Hu, Chen, & Nair, 2022). Based on this fact, we propose an iterative way to approximate the Shapley values with low-order results. We start from computing Shapley values assuming order = 1, then keep increasing the order until the Shapley values converge. In summary, our contributions are:

a. We demonstrate an additive property when computing SHAP for SHAP definitions that satisfy our additivity and dummy assumptions, including B-SHAP and kernel SHAP. With this property, SHAP can be computed very efficiently when we know the functional decomposition of the model and the components are low order.
b. We derive formulas of computing SHAP exactly for models with unknown functional decomposition but known model order. Using these formulas, SHAP values can be computed efficiently in polynomial time.
c. We also propose an iterative way to approximate SHAP with low-order formulas in (b) when the model order is unknown.
d. We show the advantage of our methods compared to sampling approach through simulations.

The remainder of the paper is organized as follows. In Section 2, we briefly review the concept of Shapley values and some existing ways of computing SHAP. We then introduce our proposed approaches. In Section 3, we show simulation results that compare speed and accuracy of different methods. We conclude our work and propose several future directions in Section 4.

---

[2] https://captum.ai/api/shapley_value_sampling.html

## 2 Methodologies

In this section, we first introduce the concept of Shapley values and some of its variants. Then we will show our proposed methods that can compute SHAP exactly and efficiently.

### 2.1 Shapley values

We let function $f(x)$ be a fitted machine-learning model that takes a p-dimensional vector $x$ as input. The Shapley value computes attributions $\phi_i$ for each feature $i$. The Shapley value of feature $i$ is given by

$$\phi_i = \sum_{u \subseteq M \setminus i} \frac{|u|!(p-|u|-1)!}{p!} \big(c(f, u+i) - c(f, u)\big), \qquad (1)$$

where $M$ is the set of all $p$ features $\{1, 2, \dots, p\}$, $c(f, u)$ is the cost function of model $f$ evaluating at set $u$. We will introduce more details about the cost function with examples in later parts. The term $c(f, u+i) - c(f, u)$ can be viewed as the contribution of feature/player $i$ given set/players $u$, and is called "gradient". An intuitive way to explain Shapley value is to compute the gradient $\big(c(f, u+i) - c(f, u)\big)$ with all subsets of $u$ and then take the weighted average. An alternative way to express Shapley value is based on permutations,

$$\phi_i = \sum_{S \subseteq Perm(p)} \frac{1}{p!} \big(c(f, Pre^i(S) + i) - c(f, Pre^i(S))\big), \qquad (2)$$

where $Perm(p)$ includes all permutations of numbers $\{1, \dots, p\}$, and $S$ is one single permutation, e.g., $S = \{3,5,1,\dots\}$. $Pre^i(S)$ is the set of predecessors of $i$ in $S$. When $i = 1$ and $S = \{3,5,1,\dots\}$, $Pre^1(S) = \{3, 5\}$. In this expression, $\phi_i$ can be viewed as the average of the gradient $\big(c(f, Pre^i(S) + i) - c(f, Pre^i(S))\big)$. This expression is widely used in computing Shapley values approximately. Castro & Gomez (2008) and Štrumbelj & Kononenko (2014) proposed methods that sample $m$ permutations among all $p!$ permutations, then compute the gradients for these sampled permutations and take the average. This estimate is unbiased because Equation (2) is the average over all permutations.

Shapley values satisfy some desirable properties: efficiency, symmetry, dummy and additivity. These properties guarantee a fair payout and are important to satisfy when computing Shapley values for different applications. A detailed discussion can be found in Section 2.6 of Sundararajan & Najmi (2020).

There are different Shapley value definitions based on different cost functions, and we list two of them. The first variant is kernel SHAP, which defines

$$c(f, u) = \int f(x) dP(x_{\bar{u}}), \bar{u} = \{1, \dots, p\} \setminus u.$$

$c(f, u)$ in kernel SHAP can be viewed as computing the average of model function value $f(x)$ over the distribution of $x_{\bar{u}}$. The second variant is Baseline SHAP (B-SHAP). In B-SHAP, we have a baseline/reference point $z$. The goal now is to explain the difference $f(x) - f(z)$ by computing the contribution of each feature. Given the baseline point $z$, the cost function in B-SHAP is defined as

$$c(f, u) = f(x_u, z_{\bar{u}}).$$

B-SHAP is a useful tool in credit lending. When a financial institution declines an application for credit, B-SHAP can be used to help explain the reason of the decision. A detailed discussion can be found in Nair et al. (2022).

There are other SHAP definitions using different cost functions. In this work, we assume that the cost function satisfies two properties.

**Assumption 1**. a. Additivity: $c(f_1 + f_2, u) = c(f_1, u) + c(f_2, u)$. b. Dummy: If $f(x) = f_v(x_v)$ only depends on $x_v$, $c(f_v, u) = c(f_v, u \cap v)$.

Additivity implies that the cost of the sum is the sum of the cost values. Dummy implies that the features that are not included in the model function won't affect the cost function. Both kernel SHAP and B-SHAP satisfy Assumption 1. However, the dummy assumption does not hold for conditional SHAP with cost function $c(f, u) = \int f(x) dP(x_{\bar{u}}|x_u)$. A more detailed discussion on conditional SHAP can be found in Chen, Janizek, Lundberg, & Lee (2020).

## 2.2 Computing SHAP with known functional decomposition

In this section, we illustrate how SHAP can be computed efficiently and exactly with known functional decomposition. We first briefly introduce functional decomposition. Many machine learning models act as a "black box": taking a high-dimensional vector as an input and predicting an output. Functional decomposition is an interpretation technique that decomposes the high-dimensional function and express it as the sum of individual feature effects and interaction effects with relatively low-dimensions. Any function can be decomposed in numerous ways. One well-established way is called functional ANOVA (fANOVA) decomposition (Hooker, 2004 and 2007), which imposes hierarchical orthogonality constraints among its components to guarantee a unique decomposition, ie, $\int f_u(x_u) f_v(x_v) w(x) dx = 0, \forall u \subset v$ and $w(x)$ is the joint distribution of predictors. Our Proposition 1 below applies to any decomposition (not only fANOVA decomposition). It proposes to compute SHAP efficiently by using the additive property of SHAP.

**Proposition 1**. Assume that the model can be decomposed into lower order terms as $f(x) = \sum_{v \subseteq M} f_v(x_v)$ and Assumption 1 holds, then $\phi_i = \sum_{i \in v} \phi_i(f_v)$, where $\phi_i(f_v)$ is the SHAP value of feature $i$ for function component $f_v$.

Intuitively, the additivity in Proposition 1 means that the SHAP value for the high-dimensional function $f(x)$ is simply the sum of SHAP value for all its low dimensional components which contain the variable $x_i$. For illustration, consider the following two simple examples.

**Example 1**. For an additive model $f(x) = \sum_j f_j(x_j)$, Proposition 1 means $\phi_i = \phi_i(f_i)$, which is $f_i(x_i) - f_i(z_i)$ in the B-shap case and $f_i(x_i) - \int f_i(x_i) dx_i$ in the kernal SHAP case.

**Example 2**. For an additive model with one two-way interaction, eg $f(x) = \sum_j f_j(x_j) + f_{12}(x_1 x_2)$, Proposition 1 states that $\phi_1 = \phi_1(f_1) + \phi_1(f_{12}), \phi_2 = \phi_2(f_2) + \phi_2(f_{12})$ and $\phi_i = \phi_i(f_i)$ for $i = 3, 4, \ldots, p$. Applying Equation (1) to the case of B-shap, $\phi_1(f_{12}) = \frac{1}{2}[f_{12}(x_1, z_2) - f_{12}(z_1, z_2)] + \frac{1}{2}[f_{12}(x_1, x_2) - f_{12}(z_1, x_2)]$. Similarly for $\phi_2(f_{12})$.

Computing $\phi_i(f_v)$ using Equation (1) is feasible when $|v|$ is small. It only involves evaluating the cost function $c(f_v, u)$ on $2^{|v|}$ subsets (eg, $2^5 = 32$). The total cost of computing $\phi_i$, $\sum_{i \in v} O(2^{|v|})$, does not grow exponentially with $p$; it is upper bounded by the order of model and number of function components. When the order of model is not high and the interactions are sparse, computing SHAP values in this way can be very efficient.

**Corollary 1**. If the variables are independent and $f(x) = \sum_{v \subseteq M} f_v(x_v)$ is the functional ANOVA decomposition which satisfies the orthogonality constraint, then for kernel SHAP, we have $\phi_i = \sum_{i \in v} \frac{1}{|v|} f_v(x_v)$.

Corollary 1 gives an explicit and simple expression for kernel SHAP under independence assumption. It attributes the prediction of each component $f_v(x_v)$ evenly among the $|v|$ variables. For example, consider the model $f(x) = \sum_j x_j + x_1 x_2$ where $x_i \sim iid$ Normal $(0,1)$. The decomposition already satisfies the orthogonality constraint because $E(x_1 x_2 \times x_1) = 0$ and $E(x_1 x_2 \times x_2) = 0$. So Corollary 1 implies that the kernel SHAP $\phi_1 = x_1 + \frac{1}{2} x_1 x_2$ and $\phi_2 = x_2 + \frac{1}{2} x_1 x_2$. However, when the variables are not independent, there is no such simple solution.

## 2.3 Computing SHAP with known order of model

In the case that we do not have the functional decomposition of the model, we can still compute SHAP efficiently in polynomial time if we know the order of the model. Theorem 1 below does not require to know the decomposition; it relies only on the overall model prediction $f(x)$ and the order of model.

**Theorem 1.** Assume the model $f(x)$ has order $K$ and Assumption 1 holds, then,

(1) When the model is additive, i.e., $K = 1$, we have for any subset $u \subseteq M \setminus i$,

$$\phi_i = c(f, u + i) - c(f, u).$$

(2) When the model has up to 2-way interactions, i.e., $K = 2$, we have for any subset $u \subseteq M \setminus i$,

$$\phi_i = \frac{1}{2}\big(c(f, u + i) - c(f, u) + c(f, \bar{u}) - c(f, \overline{u + \iota})\big), \bar{u} = M \setminus u.$$

In other words, $\phi_i$ is the average of gradients w.r.t. (any) set $u$ and its complement set $\overline{u + \iota}$.

(3) When $K \geq 3$, let $q = \lfloor (K-1)/2 \rfloor$, where $\lfloor \ \rfloor$ is the floor function. Then,

$$\phi_i = \sum_{m=0}^{q} a_m (d_m + d_{p-m-1}),$$

where $d_m = \frac{1}{\binom{p-1}{m}} \sum_{u:|u|=m, u \subseteq M \setminus i} \big(c(f, u + i) - c(f, u)\big)$ is the average gradient for all subsets with cardinality $m$, and the coefficients $a_m$ can be obtained by solving the $q+1$ equations

$$2 \sum_{m=r}^{q} a_m \frac{\binom{p-2r-1}{m-r}}{\binom{p-1}{m}} = \frac{r! r!}{(2r+1)!}, r = 0, \ldots, q \quad (3)$$

The complete proof is in Appendix and here we provide a brief explanation focusing on B-shap. In Example 1, we know $\phi_i = f_i(x_i) - f_i(z_i)$ from previous discussion. On the other hand, for any subset $u \subseteq M \setminus i$, $c(f, u+i) - c(f, u) = \big[\sum_{j \in u+i} f_j(x_j) + \sum_{j \notin u+i} f_j(z_j)\big] - \big[\sum_{j \in u} f_j(x_j) + \sum_{j \notin u} f_j(z_j)\big] = f_i(x_i) - f_i(z_i)$. So the conclusion holds and we only need to evaluate the cost function twice. In Example 2, $\phi_1 = f_1(x_1) - f_1(z_1) + \frac{1}{2}[f_{12}(x_1, z_2) - f_{12}(z_1, z_2)] + \frac{1}{2}[f_{12}(x_1, x_2) - f_{12}(z_1, x_2)]$ from previous discussion. On the other hand, $c(f, u+1) - c(f, u) = f_1(x_1) - f_1(z_1) + \begin{cases} f_{12}(x_1, x_2) - f_{12}(z_1, x_2), & \text{if } 2 \in u \\ f_{12}(x_1, z_2) - f_{12}(z_1, z_2), & \text{if } 2 \notin u \end{cases}$. The gradient depends only on if $2 \in u$ or not, and the true $\phi_1$ is an average of the two scenarios. So a natural choice is to select any subset $u$ and its complement $\overline{u+1}$ and take average on the two gradients. This approach works for any order-2 models. The computation cost is only 4 cost function evaluations.

When the model is of order $K \geq 3$, given $K$ is relatively small compared to $p$, it may suggest us to only look at cost functions for subsets with small cardinalities. However, as seen from the order-2 case, we also need the complement subsets to "balance out" to get an unbiased estimate. Therefore, we will also include the subsets with high cardinalities. The final step is to find the appropriate coefficients $a_m$'s. This is done by solving Equation (3). Note Equations (3) forms a triangular system of linear equations which can be solved easily and the solution uniquely exists.

Using the formula in Theorem 1 can improve the speed of computing SHAP significantly, especially when the model order is not high, because only the subsets on two "tails" will be evaluated. Imagine the case that the model order is 3 or 4, then $q = 1$. So we only need the subsets with cardinality 0, 1, $p-1$, $p-2$. The total number of function evaluations is $2(1 + (p-1)) = 2p$. When the order is 5 or 6, then $q = 2$, we only need the subsets with cardinality $0, 1, 2, p-1, p-2$ and $p-3$, and the total number of function evaluations is $2\left(1 + (p-1) + \binom{p-1}{2}\right) = O(p^2)$. This linear and quadratic time complexity is much faster than the exponential time consuming in Equation (1) and (2).

## 2.4 Approximating SHAP with unknown order of model

Using the order-K formula in Theorem 1 can be efficient to compute B-SHAP exactly when the model order is known and not high. In practice, however, we might not know the order of a model, or the order can be high but the high order interactions are small. In this case, we can approximate the SHAP values using an iterative way. Specifically, we start from computing SHAP with K=1. Then we increase the order to 2 and compute the difference between SHAP of order 2 and order 1. We keep increasing the order by 2 (since the formula is the same for K and K-1 when K is even and K ≥ 4) until the difference between SHAP of order K and SHAP of order K-2 are small enough. Then we use the order-K result as our estimated SHAP. The reasoning behind this procedure is that if the model is approximately low-order, then the SHAP values won't change much when K reaches certain sufficiently large value. Note that using a higher order formula for a lower order model won't cause any error, so the results of order $K_{true} + 2$ and order $K_{true}$ are exactly the same, and the algorithm will always converge at order $K_{true} + 2$ or earlier. This procedure is summarized in Table 2-1.

**Table 2-1. Iterative way to compute SHAP**

| |
|---|
| Input: data to be explained, max_order, threshold |
| current_order = 1, difference = positive infinity, converge = False <br> while current_order ≤ max_order and difference > threshold: <br>     difference = mean(\|SHAP of current order - SHAP of previous order\|)$^2$/Variance(SHAP of current order) <br>     if difference < threshold: <br>         converge = True <br>         return SHAP results, current_order and converge <br>     if current_order = 1: <br>         current_order += 1 <br>     else: <br>         current_order += 2 <br> return SHAP results, current_order and converge |
| Output: SHAP results, order at convergence or max order, convergence (Boolean) |

Note during the iterative procedure, it is possible that the SHAP values for most features and observations converge, but they do not converge on a few features for a few observations. The reason can be that these features might be involved in the high order interactions and the magnitude of the high order terms are large for these observations. In that case, the algorithm can be further optimized to drop observations/features which have converged and focus on the ones which haven't.

## 3 Simulations

In this section, we compare different methods to compute SHAP in term of speed and accuracy of these approaches. We first show the simulation settings and then show our results. Although our proposed methods can be applied to different types of SHAP, our simulations focus on B-SHAP.

### 3.1 Simulation settings

In our simulations, we consider models of order 2, 4 and 6. We compare Captum with our exact methods using order-K and functional decomposition formular, to see how much error the Captum method has. We also examine for the order-6 model, how well the lower order approximation method works. Below are the models we consider:

Order-2 model: $\sum_{j=1}^{10} x_j + x_1 x_2 + x_3 x_4 + x_5 x_6 + x_7 x_8$

Order-4 model: $\sum_{j=1}^{10} x_j + x_1 x_2 + x_3 x_4 + x_5 x_6 + x_7 x_8 + x_1 x_2 x_3 x_4 + x_5 x_6 x_7 x_8$

Order-6 model: $\sum_{j=1}^{10} x_j + x_1 x_2 + x_3 x_4 + x_5 x_6 + x_7 x_8 + x_1 x_2 x_3 x_4 + x_5 x_6 x_7 x_8 + \alpha x_1 x_2 x_3 x_4 x_5 x_6$

For the order-6 model, the coefficients for the 6-way interaction term α are 0.5, 1 and 2 to reflect different levels of high-order interaction effect. The input variables are i.i.d. normal with mean 0 and variance 1 and we generated 10000 observations. Since our goal here is not model fitting, we simply use the true models and compute B-SHAP for the true models. When computing B-SHAP, we use all 10000 data points and compare them to two baseline points. The first choice of baseline is the average of the 10000 data points, and the second choice is the 97.5 percentile of each feature. The averages are close to 0 and the 97.5 percentiles are close to 1.96 because each feature is generated with a standard normal distribution.

### 3.2 Simulation results

#### 3.2.1 Accuracy comparison

In this section, we first compare the accuracy of B-SHAP computed using the sampling method in Captum with the exact B-SHAP computed using order-K and functional decomposition (f-dcmp) formulas. We use 25 and 100 as the number of samples (subsets) in Captum. Figure 3-1 and Figure 3-2 show the comparison for order-2 and order-4 model respectively. The top plots are for the average baseline and the bottom figures are for the 97.5 percentile baseline. The left plots show the results with 25 samples and the right ones are for the 100 samples. Because SHAP values for main-effect only variables can always be estimated correctly, we don't include $\phi_9$ and $\phi_{10}$ in the figures. We can see that the sampling method does not perform well when the number of samples is 25 as some points are far from the true values. Note for this particular sample of 25 subsets, $\phi_5$ and $\phi_6$ are estimated with high accuracy for the order-2 model. This happens purely by chance; a change of random seed in Captum will result in $\phi_5$ and $\phi_6$ being estimated less accurately. When the number of samples increases to 100, the performance becomes better, yet small errors still exist. Moreover, the scale of error becomes larger when the baseline point is the 97.5 percentile. To explain this for the order-2 model, we first compute the error related with Captum sampling approach. Suppose we are interested in SHAP value for $x_1$. For any permutation $S$, we have

$$c(f, Pre^1(S) + 1) - c(f, Pre^1(S)) = \begin{cases} (x_1 - z_1)(1 + x_2), if\ 2 \in Pre^1(S) \\ (x_1 - z_1)(1 + z_2), if\ 2 \notin Pre^1(S) \end{cases}.$$

Since the events $2 \in Pre^1(S)$ and $2 \notin Pre^1(S)$ have equal probability, we have $E(\hat{\phi}_1) = (x_1 - z_1)\left(1 + \frac{x_2+z_2}{2}\right) = \phi_1$, which is unbiased. The standard error is $Std(\hat{\phi}_1) = \frac{|(x_1-z_1)(x_2-z_2)|}{\sqrt{4m}}$, where $m$ is the number of samples. The numerator, $|(x_1 - z_1)(x_2 - z_2)|$, measures the strength of $x_1 x_2$-interaction from the sample $x$ to the reference point $z$. When $(z_1, z_2)$ changes from the dense center $(0,0)$ to the far extreme of 97.5 percentile $(1.96, 1.96)$, the interaction gets stronger for most observations, hence the overall standard error increases. The explanation for the order-4 model is similar but more cumbersome, so we omit it here. Note when we happen to have equal number of permutations between $2 \in Pre^1(S)$ and $2 \notin Pre^1(S)$, $\phi_1$ can be estimated precisely. This explains the high accuracy for $\hat{\phi}_5$ and $\hat{\phi}_6$ we see in Figure 3-1.

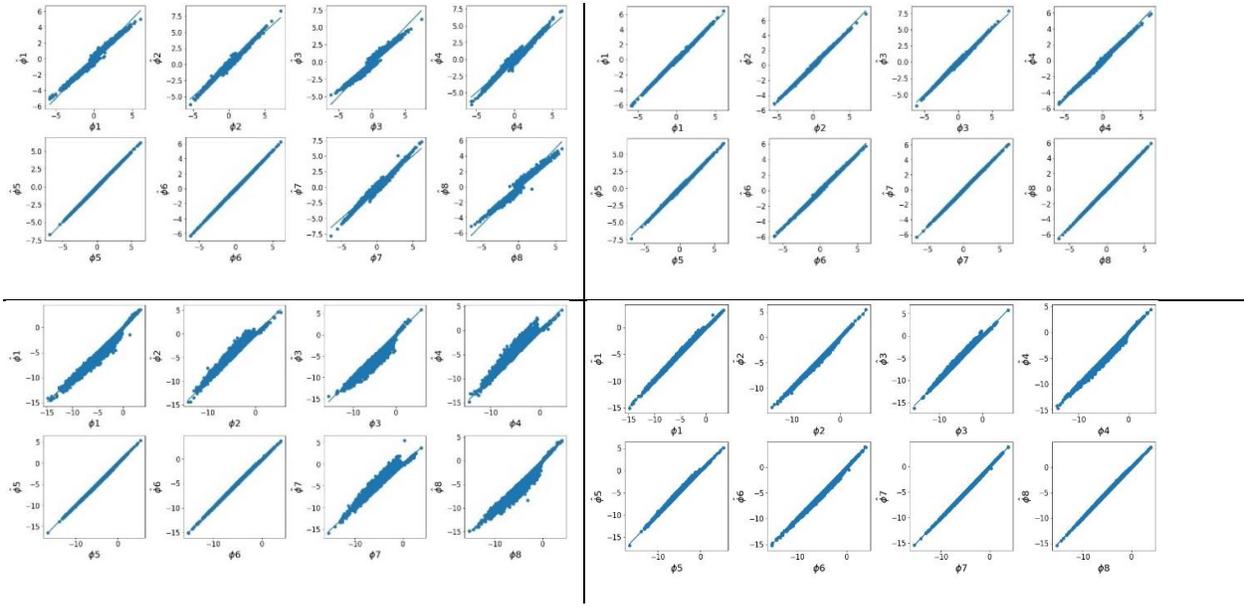

**Figure 3-1.** Computing B-SHAP for order-2 model using sampling method in Captum. We compared the sampled B-SHAP $\hat{\phi}_i$ with true $\phi_i$. The top figures use the average as the baseline while the bottom ones have 97.5 percentile as the baseline point. The left figures use 25 samples, the right figures use 100 samples.

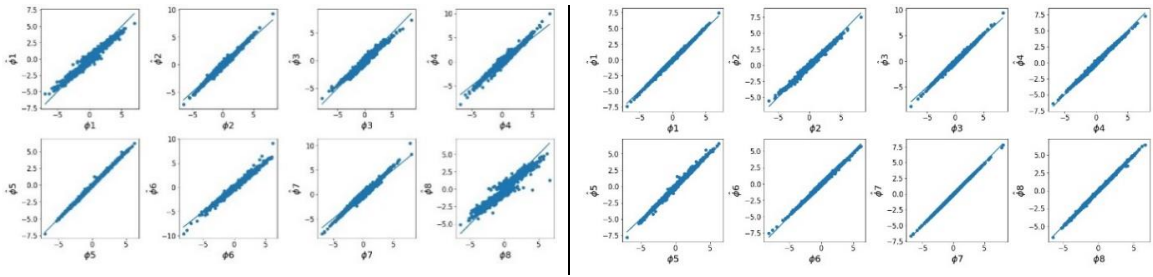

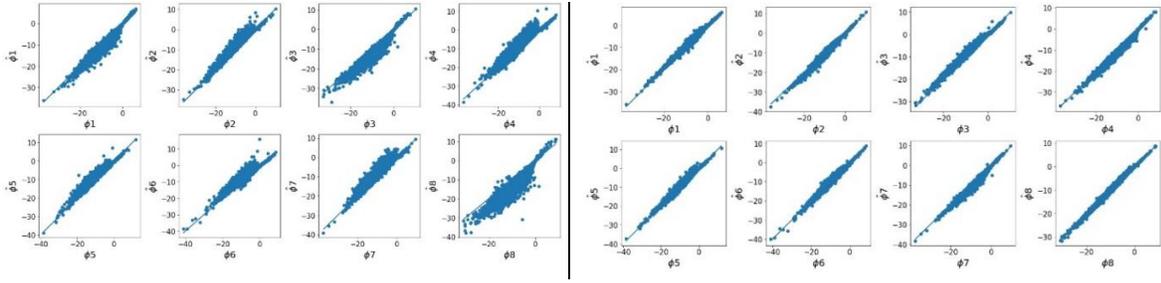

**Figure 3-2.** Computing B-SHAP for order-4 model using sampling method in Captum. The top figures use the average as the baseline while the bottom ones have 97.5 percentile as the baseline point. The left figures use 25 samples, the right figures use 100 samples.

For the order-6 model, we choose different coefficients 0.5, 1 and 2 for the 6-way interaction. We still run the sampling method with number of samples 25 and 100. We then run the iterative method proposed in Section 2.4. We set the max order in the iterative method as 10 because in practice it is rare to see interactions higher than that. Recall we use the relative difference to check convergence, where the relative difference is defined as mean(|SHAP of current order - SHAP of previous order|)$^2$/Variance(SHAP of current order). We set the threshold for convergence to be 0.0001. When the coefficient is 0.5 and the baseline is the average, the iterative approach stops at order = 6 because the relative difference between order-4 and order-6 SHAP values is smaller than the threshold. This indicates the 6-way interaction is weak. When the coefficient is 0.5 and the baseline is the 97.5 percentile, or when the coefficient is 1 or 2 regardless of baseline, the strength of 6-way interaction is stronger. Therefore, the 4$^{th}$-order approximation has larger errors, and the iterative approach stops at order = 8, meaning that the difference between order-4 approximation and order-6 approximation is larger than threshold, and the difference between order-6 and order-8 approximations is small (0). To further demonstrate this, we show the relative difference between SHAP values with different orders and the true SHAP values in Figure 3-3. The top figure has the average as the baseline and the bottom one is with the 97.5 percentile. Since the relative difference at order 1 is much bigger than the rest (around 0.07 for the average baseline and larger than 50 for the 97.5 percentile baseline), our plots start from order 2 for better illustration. We can see that the relative differences decrease significantly when the order increases from 2 to 4 for all 3 coefficients. The relative differences are already close to 0 at order 4 when the baseline is average; however, they are much larger at order 4 when the baseline is the 97.5 percentile.

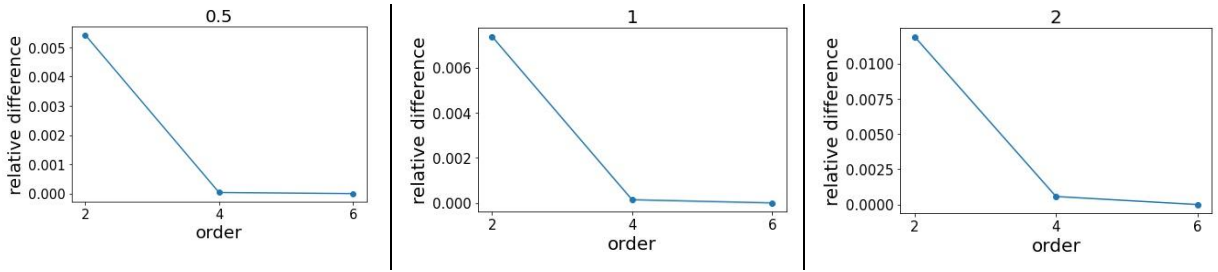

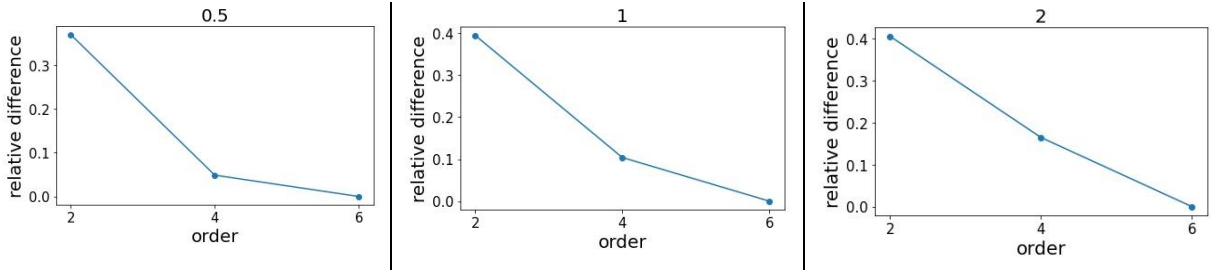

**Figure 3-3.** The relative difference between SHAP values with different orders and the true SHAP values. The top figures use the average as the baseline while the bottom ones have 97.5 percentile as the baseline point. The titles show the coefficients.

Finally, we compare the results of Captum method with our iterative algorithm. Since we set the convergence threshold to be very small (0.0001), our iterative algorithm stops at either order 6 or 8, both yielding the exact result. Figure 3-4, 3-5, 3-6 show the comparison of Captum method with our exact result for coefficient 0.5, 1 and 2 respectively. The top figures are for the average baseline and the bottom plots have the 97.5 percentile as baseline. The left plots are with 25 samples and the right ones are with 100 samples. As we explained before, the interaction strength gets stronger when the baseline is the 97.5 percentile. Hence the errors in bottom plots are much larger than the top ones. Larger coefficients of high-way interactions also increase the interaction strength, thus we can see larger errors in Figures 3-5 and 3-6 than Figure 3-4.

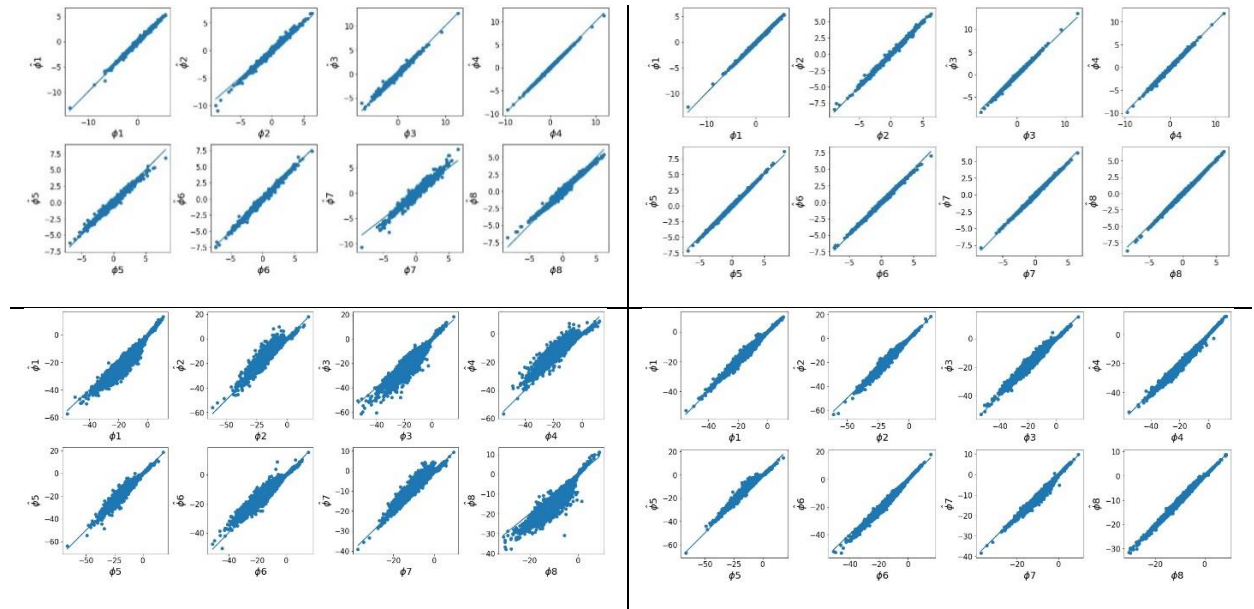

**Figure 3-4.** Computing B-SHAP for order-6 model with coefficient 0.5 using sampling method in Captum. We compared the sampled B-SHAP $\widehat{\phi}_i$ with true $\phi_i$. The top figures use the average as the baseline while the bottom ones have 97.5 percentile as the baseline point. The left figures use 25 samples, the right figures use 100 samples.

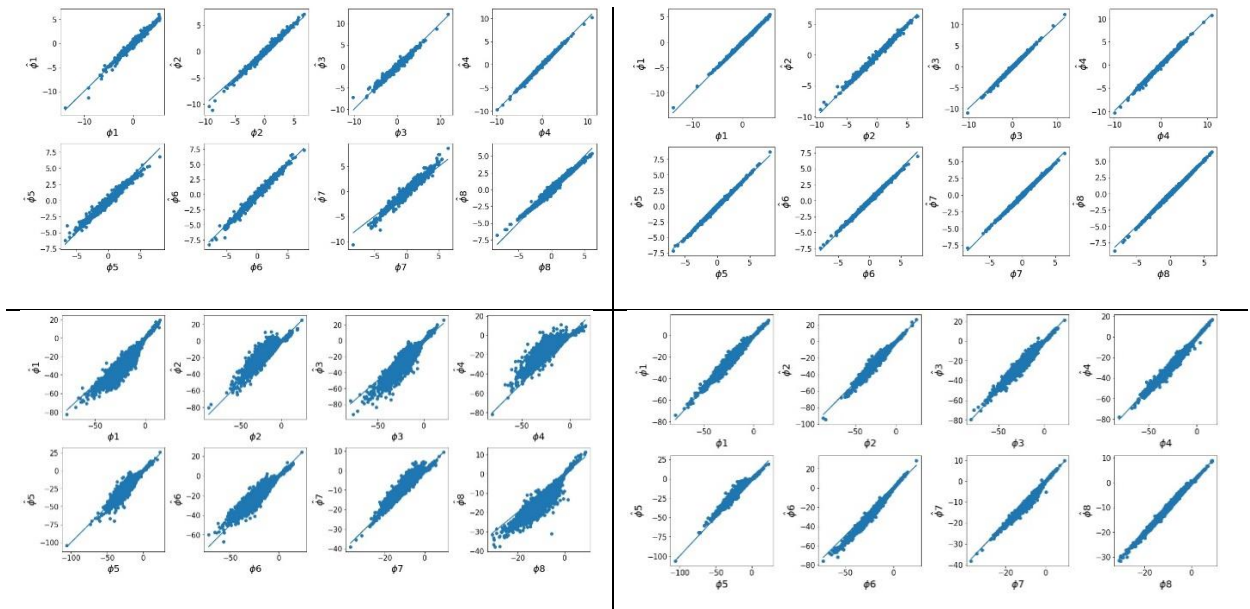

**Figure 3-5.** Computing B-SHAP for order-6 model with coefficient 1 using sampling method in Captum. We compared the sampled B-SHAP $\widehat{\phi}_i$ with true $\phi_i$. The top figures use the average as the baseline while the bottom ones have 97.5 percentile as the baseline point. The left figures use 25 samples, the right figures use 100 samples.

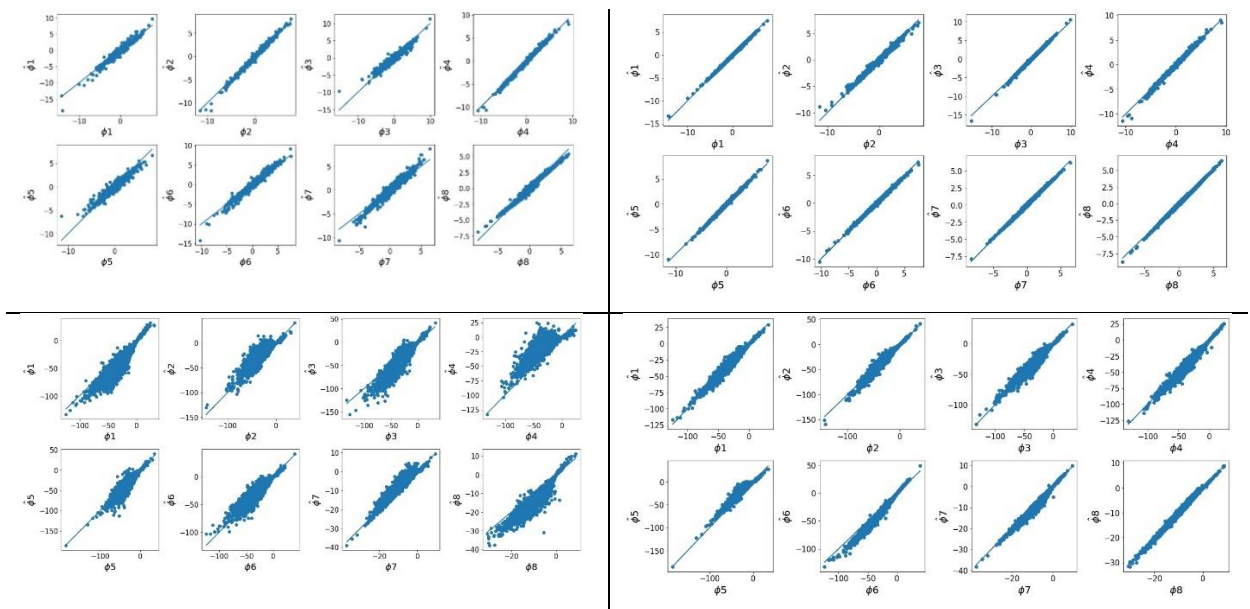

**Figure 3-6.** Computing B-SHAP for order-6 model with coefficient 2 using sampling method in Captum. We compared the sampled B-SHAP $\widehat{\phi}_i$ with true $\phi_i$. The top figures use the average as the baseline while the bottom ones have 97.5 percentile as the baseline point. The left figures use 25 samples, the right figures use 100 samples.

### 3.2.2 Speed comparison

In this section, we show the speed of different methods of computing SHAP. Note speed is not only related with the method but also the numeric implementation. We didn't apply any high-performance computing tools here so the speed should be considered as the baseline not the best scenario. We use the 10-variable models specified in Section 3.1 as well as three new models each with 10 more variables, e.g., the order-2 model with 20 variables now becomes $\sum_{j=1}^{20} x_j + x_1x_2 + x_3x_4 + x_5x_6 + x_7x_8$. As before, we use all 10000 observations to compute B-SHAP. We show the results of using the average as the baseline in the speed comparison. The speeds of using the 97.5 percentile as baseline are very close to those of the average baseline; the only difference with average baseline is that when the model has 6-way interaction with coefficient 0.5, the iterative method stops at order 6 instead of 8. When applying the f-dcmp method, we assume that we know the model form, hence we compute the B-SHAP of each component using Equation (1) and then add them up following Proposition 1. When using the order-K formula, we use the order of the true model. We use 25 and 100 as the sample sizes for Captum. For order-6 models, we also apply the iterative method. The max order is 10 and the threshold for the relative difference is 0.0001.

Table 3-1 and 3-2 show the time of running each method on the same computing platform with 10 and 20 variables respectively. We first notice that applying f-dcmp is fastest for almost all the models. Applying the order-K formula is faster than Captum-25 when the order is low and number of variables is small. When order is 6 and number of variables is 20, the order-K formula is slower than Captum-25 and close to Captum-100. Note that applying the order-K formula gives the exact results. Therefore, if we know the order of the model, using the order-K formula is a better choice than the sampling method in both accuracy and speed when the order is low, and the number of variables is not large. In the case that we do not know the order of the model and the high-order effect is small, e.g., with coefficient 0.5, we see that applying the iterative approach can be a good strategy. It is still faster than Captum-100 when there are 10 variables and as fast as Captum-100 for 20 variables. With strong high-order effect (coefficient 1 and 2), the iterative procedure stops at order 8, hence the time is longer than Captum-100.

**Table 3-1. Time comparison for each method when the models have 10 variables. The time is in seconds.**

|  | Order 2 | Order 4 | Order 6 – 0.5 | Order 6 – 1 | Order 6 – 2 |
|---|---|---|---|---|---|
| f-dcmp | 0.066 | 0.17 | 0.5 | 0.48 | 0.52 |
| order-K | 0.034 | 0.17 | 0.63 | 0.64 | 0.6 |
| Iterative |  |  | 0.59 | 1.33 | 1.24 |
| Captum-25 | 0.57 | 0.56 | 0.59 | 0.59 | 0.58 |
| Captum-100 | 2.35 | 2.18 | 2.34 | 2.25 | 2.39 |

**Table 3-2. Time comparison for each method when the models have 20 variables. The time is in seconds.**

|  | Order 2 | Order 4 | Order 6 – 0.5 | Order 6 – 1 | Order 6 – 2 |
|---|---|---|---|---|---|
| f-dcmp | 0.1 | 0.21 | 0.54 | 0.53 | 0.5 |
| order-K | 0.078 | 0.84 | 5.62 | 5.85 | 5.69 |
| Iterative |  |  | 5.41 | 26.8 | 25.28 |
| Captum-25 | 1.14 | 1.17 | 1.24 | 1.23 | 1.32 |
| Captum-100 | 4.61 | 4.6 | 5.01 | 5.04 | 5.17 |

## 4 Future works

In this work, we propose three ways of computing SHAP values efficiently. The functional decomposition approach is both fast and accurate if we can express the model as the sum of low-order components, for example, as in the fANOVA models. In general, it can be challenging to obtain such decompositions. The second approach is to apply the order-K formula if we know the order of model, as in xgboost model with maximum depth K. This reduces the computation complexity from exponential to polynomial and is also efficient when K is small. Finally, if we don't know anything about the model, we propose the iterative algorithm. This approach can be still computationally expensive when the number of features is large and order of interactions high. One possible solution is to improve our iterative algorithm to drop observations/features which have converged and only focus on the ones which haven't.

## Appendix

### Proof of Proposition 1

Given additivity assumption, we have $c(f, u) = c(\sum_v f_v, u) = \sum_v c(f_v, u) = \sum_v c(f_v, u \cap v)$. Therefore, $\phi_i = \sum_{u \subseteq M \setminus i} \frac{|u|!(p-|u|-1)!}{p!} \big(c(f, u+i) - c(f, u)\big) = \sum_{u \subseteq M \setminus i} \frac{|u|!(p-|u|-1)!}{p!} \big(\sum_v \big(c(f_v, u+i) - c(f_v, u)\big)\big) = \sum_v \sum_{u \subseteq M \setminus i} \frac{|u|!(p-|u|-1)!}{p!} \big(c(f_v, u+i) - c(f_v, u)\big)$. Due to dummy assumption, $c(f_v, u+i) - c(f_v, u) = \begin{cases} 0 & i \notin v \\ c(f_v, u \cap v + i) - c(f_v, u \cap v) & i \in v \end{cases}$. This means (1) we only need to look at $f_v$ for which $i \in v$; (2) the gradients are the same for all subsets $u$ which has the same $u \cap v$. Let $u' = u \cap v$, we need to collect coefficients on the same gradient term, i.e., $\sum_{u:u\subseteq M\setminus i, u\cap v=u'} \frac{|u|!(p-|u|-1)!}{p!} = \sum_{|u|=|u'|}^{p-1} \frac{|u|!(p-|u|-1)!}{p!} \binom{p-|v|}{|u|-|u'|} = \frac{(p-|v|)!|u'|!(|v|-|u'|-1)!}{p!} \sum_{|u|=|u'|}^{p-1} \binom{|u|}{|u'|}\binom{p-|u|-1}{|v|-|u'|-1}$.

By Vandermonde identity, the summation $\sum_{|u|=|u'|}^{p-1} \binom{|u|}{|u'|}\binom{p-|u|-1}{|v|-|u'|-1} = \binom{p}{|v|}$, so the coefficient simplifies to $\frac{|u'|!(|v|-|u'|-1)!}{|v|!}$. Therefore, $\phi_i = \sum_{v:i\in v}\sum_{u'\subseteq v\setminus i} \frac{|u'|!(|v|-|u'|-1)!}{|v|!} (c(f_v, u'+i) - c(f_v, u')) = \sum_{v:i\in v} \phi_i(f_v)$.

**Proof of Corollary 1**

We only need to show that $\phi_i(f_v) = \frac{1}{|v|} f_v(x_v)$. Given the orthogonality property in fANOVA decomposition, we have $c(f_v, u) = \int f_v(x_v) dx_{v\setminus u} = \begin{cases} f_v(x_v), & \text{if } u = v \\ 0, & \text{if } u \subset v \end{cases}$. Thus $c(f_v, u+i) - c(f_v, u) = \begin{cases} f_v(x_v), & \text{if } u = v\setminus i \\ 0, & \text{if } u \subset v\setminus i \end{cases}$. This leads to $\phi_i(f_v) = \sum_{u\subseteq v\setminus i} \frac{|u|!(|v|-|u|-1)!}{|v|!}(c(f_v, u+i) - c(f_v, u)) = \frac{|v\setminus i|!(|v|-|v\setminus i|-1)!}{|v|!} f_v(x_v) + 0 = \frac{1}{|v|} f_v(x_v)$.

**Proof of Theorem 1**

When the model is additive, $f = \sum_{i=1}^p f_i$. By Proposition 1, $\phi_i = \phi_i(f_i) = c(f_i, i) - c(f_i, \emptyset)$. On the other hand, by Assumption 1, $c(f, u+i) - c(f, u) = c(f_i, i) - c(f_i, \emptyset)$ for any subset $u$. Therefore, $\phi_i = c(f, u+i) - c(f, u)$.

When the model has up to two-way interactions, $f = \sum_{i=1}^p f_i + \sum_{i<j} f_{ij}$. By Proposition 1, $\phi_i = \phi_i(f_i) + \sum_j \phi_i(f_{ij}) = c(f_i, i) - c(f_i, \emptyset) + \sum_j \frac{1}{2}\left(c(f_{ij}, i) - c(f_{ij}, \emptyset) + c(f_{ij}, i+j) - c(f_{ij}, j)\right)$. On the other hand, by Assumption 1, $c(f, u+i) - c(f, u) = c(f_i, i) - c(f_i, \emptyset) + \sum_j \left(c(f_{ij}, u\cap j + i) - c(f_{ij}, u\cap j)\right) = c(f_i, i) - c(f_i, \emptyset) + \sum_{j\in u}\left(c(f_{ij}, j+i) - c(f_{ij}, j)\right) + \sum_{j\in \overline{u+i}}\left(c(f_{ij}, i) - c(f_{ij}, \emptyset)\right)$. Replacing $u$ with $\overline{u+i}$ in above identity, we can get $c(f, \bar{u}) - c(f, \overline{u+i}) = c(f_i, i) - c(f_i, \emptyset) + \sum_{j\in \overline{u+i}}\left(c(f_{ij}, j+i) - c(f_{ij}, j)\right) + \sum_{j\in u}\left(c(f_{ij}, i) - c(f_{ij}, \emptyset)\right)$. Now it is easy to see $\phi_i = \frac{1}{2}\left(c(f, u+i) - c(f, u) + c(f, \bar{u}) - c(f, \overline{u+i})\right)$.

For models with three-way or higher order interactions, by Proposition 1, $\phi_i = \sum_{v:i\in v}\sum_{u'\subseteq v\setminus i} \frac{|u'|!(|v|-|u'|-1)!}{|v|!}(c(f_v, u'+i) - c(f_v, u'))$. On the other hand, by Assumption 1, $d_m = \frac{1}{\binom{p-1}{m}} \sum_{u:|u|=m}(c(f, u+i) - c(f, u)) = \frac{1}{\binom{p-1}{m}} \sum_{u:|u|=m}\sum_{v:i\in v}(c(f_v, u'+i) - c(f_v, u')) = \sum_{v:i\in v}\sum_{u'\subseteq v\setminus i} \frac{\binom{p-|v|}{m-|u'|}}{\binom{p-1}{m}}(c(f_v, u'+i) - c(f_v, u'))$, where $u' = u\cap v$ and the last equation holds because there are $\binom{p-|v|}{m-|u'|}$ number of $u$ sets which satisfy $|u| = m$ and $u\cap v = u'$. Thus to prove $\phi_i = \sum_{m=0}^q a_m(d_m + d_{p-m-1})$, we need to make sure the coefficients on $c(f_v, u'+i) - c(f_v, u')$ are the same on both sides of equation. Let $k = |v|$ and $j = |u'|$, we need to prove the following equation under the conditions of Equation (3) ($j < k$ since $u' \subseteq v \setminus i$):

$$\sum_{m=0}^{q} a_m \frac{\binom{p-k}{m-j}+\binom{p-k}{p-m-j-1}}{\binom{p-1}{m}} = \frac{j!(k-j-1)!}{k!}, 0 \leq j < k \leq K \tag{4}$$

Plug in $r = 0$ to Equation (3) we get $2\sum_{m=0}^{q} a_m = 1$. This means Equation (4) hold for $(j, k) = (0,1)$. For $(j, k) = (0,2)$, the LHS of Equation (4) is $\sum_{m=0}^{q} a_m \frac{\binom{p-2}{m}+\binom{p-2}{m-1}}{\binom{p-1}{m}} = \sum_{m=0}^{q} a_m \frac{\binom{p-1}{m}}{\binom{p-1}{m}} = \frac{1}{2}$ (The first equality uses Pascal's rule). Hence Equation (4) holds for $(0, 2)$. Similarly, we can show it holds for $(1,2)$ as well.

Now assume Equation (4) holds for $k = 1, \dots, k'-1$. For $k = k'$, we consider two cases where $k'$ is even or odd.

(I) when $k' = 2t+1$ is odd, for $(j, k) = (t, 2t+1)$, Equation (4) becomes $2\sum_{m=0}^{q} a_m \frac{\binom{p-2t-1}{m-t}}{\binom{p-1}{m}} = \frac{t!t!}{(2t+1)!}$, it holds due to Equation (3) with $r = t$. Now use deduction for other pairs. Suppose Equation (4) hold for the pair $(j, k')$; for the pair $(j+1, k')$, we sum up the LHS of Equation (4) for the two pairs:

$\sum_{m=0}^{q} a_m \frac{\binom{p-k'}{m-j-1}+\binom{p-k'}{p-m-j-2}+\binom{p-k'}{m-j}+\binom{p-k'}{p-m-j-1}}{\binom{p-1}{m}} = \sum_{m=0}^{q} a_m \frac{\binom{p-k'+1}{m-j}+\binom{p-k'+1}{p-m-j-1}}{\binom{p-1}{m}} = \frac{j!(k'-j-2)!}{(k'-1)!}$, where the last equality holds by applying Equation (4) to the pair $(j, k'-1)$. Given the LHS for $(j, k')$ equals $\frac{j!(k'-j-1)!}{k'!}$, the LHS for $(j+1, k')$ equals $\frac{j!(k'-j-2)!}{(k'-1)!} - \frac{j!(k'-j-1)!}{k'!} = \frac{(j+1)!(k'-j-2)!}{k'!}$, which is the RHS of Equation (4). Hence Equation (4) holds for the pair $(j+1, k')$ as well. By induction, Equation (4) holds for all $(j, k')$, where $t \leq j \leq k'$. Since both the LHS and RHS in Equation (4) are same for $(j, k')$ and $(k'-j-1, k')$, Equation (4) holds for all $(j, k')$ pairs.

(II) when $k' = 2t$ is even, for $(j, k') = (t, 2t)$, the LHS of Equation (4) is $\sum_{m=0}^{q} a_m \frac{\binom{p-2t}{m-t}+\binom{p-2t}{p-m-t-1}}{\binom{p-1}{m}} = \sum_{m=0}^{q} a_m \frac{\binom{p-2t+1}{m-t+1}}{\binom{p-1}{m}} = \frac{(t-1)!(t-1)!}{2(2t-1)!} =$RHS of Equation (4), where the second equality comes from Equation (3) with $r = t-1$. Using the same argument in case (I), we can show if Equation (4) hold for the pair $(j, k')$, it must also hold for the pair $(j+1, k')$. Therefore, it must hold for all $(j, k')$ pairs where $t \leq j < k'$. For $j < t$, since both the LHS and RHS are same for $(j, k')$ and $(k'-j-1, k')$, Equation (4) holds for all $(j, k')$ pairs.

Now we have proved Equation (4) hold for $k = k'$. By induction, Equation (4) hold for all $0 \leq j < k \leq K$. This ends the proof. ∎